\documentclass[letterpaper]{article} 
\usepackage{aaai25}  
\usepackage{times}  
\usepackage{helvet}  
\usepackage{courier}  
\usepackage[hyphens]{url}  
\usepackage{graphicx} 
\usepackage{natbib}  
\usepackage{caption} 
\usepackage{amsfonts}
\usepackage{amsmath}
\usepackage{algorithm}
\usepackage{algorithmic}
\usepackage{multirow}
\usepackage{newfloat}
\usepackage{listings}
\usepackage{url}
\DeclareCaptionStyle{ruled}{labelfont=normalfont,labelsep=colon,strut=off} 
\floatstyle{ruled}
\newfloat{listing}{tb}{lst}{}
\floatname{listing}{Listing}
\title{What to Preserve and What to Transfer: \\ Faithful, Identity-Preserving Diffusion-based Hairstyle Transfer}
\author{
    Chaeyeon Chung, Sunghyun Park, Jeongho Kim, and Jaegul Choo
}
\affiliations{
    KAIST AI, South Korea\\
    cy\_chung@kaist.ac.kr,  psh01087@gmail.com, \{rlawjdghek, jchoo\}@kaist.ac.kr
}

\usepackage{bibentry}

\begin{document}
\twocolumn[{%
\renewcommand\twocolumn[1][]{#1}%
\maketitle
\vspace{-1cm}
\begin{center}
    \centering
    \captionsetup{type=figure}
    \includegraphics[width=\linewidth]{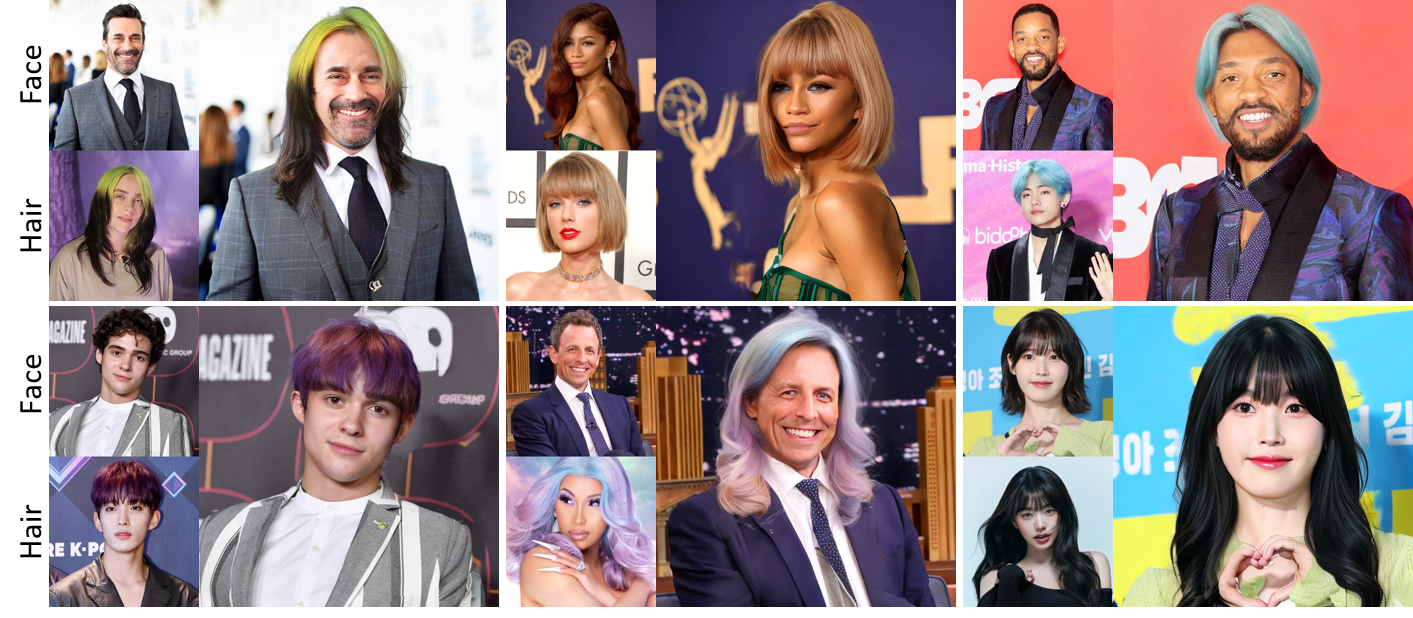}
    \vspace{-0.5cm}
    \captionof{figure}{Generated results by HairFusion using in-the-wild images. 
    Given a pair of a face image and a reference hairstyle image, our method can generate high-fidelity images.
    These results show our model's generalizability to diverse face images. 
    }
    \label{fig:teaser}
\end{center}
}]
\begin{abstract}
Hairstyle transfer is a challenging task in the image editing field that modifies the hairstyle of a given face image while preserving its other appearance and background features.
The existing hairstyle transfer approaches heavily rely on StyleGAN, which is pre-trained on cropped and aligned face images.
Hence, they struggle to generalize under challenging conditions such as extreme variations of head poses or focal lengths.
To address this issue, we propose a one-stage \textbf{hair}style transfer dif\textbf{fusion} model, HairFusion, that applies to real-world scenarios.
Specifically, we carefully design a hair-agnostic representation as the input of the model, where the original hair information is thoroughly eliminated.
Next, we introduce a hair align cross-attention (Align-CA) to accurately align the reference hairstyle with the face image while considering the difference in their head poses.
To enhance the preservation of the face image's original features, we leverage adaptive hair blending during the inference, where the output's hair regions are estimated by the cross-attention map in Align-CA and blended with non-hair areas of the face image.
Our experimental results show that our method achieves state-of-the-art performance compared to the existing methods in preserving the integrity of both the transferred hairstyle and the surrounding features.
The codes are available at \url{https://github.com/cychungg/HairFusion}.
\end{abstract}

\section{Introduction}




With the recent advancement of generative models~\cite{rombach2022high, saharia2022photorealistic}, image editing technologies~\cite{zhang2023adding, yang2023paint, hertz2022prompt} have shown impressive results.
The task of hairstyle transfer is one of the most challenging image editing tasks, focusing on modifying an input face image's hairstyle to the reference hairstyle.
This technology can offer users a preview of how different hairstyles would look on them, enhancing customer satisfaction.
The key challenge of hairstyle transfer is to transfer the reference hairstyle to the face while preserving other features (\textit{e.g.}, identity, clothing, and background) in the face image.

Recent hairstyle transfer~\cite{kim2022style, khwanmuang2023stylegan, wei2023hairclipv2, nikolaev2024hairfastgan} allows users to manipulate hairstyles using StyleGAN2~\cite{karras2020stylegan2}, which can generate high-fidelity face images.
However, such GAN-based approaches still suffer from several challenges.
First, since most of them rely on StyleGAN2 pre-trained on FFHQ dataset~\cite{karras2019stylegan}, they often fail to generalize to face images with various head poses or focal lengths that lie outside the latent space of StyleGAN2.
This hinders its application to real-world scenarios~\cite{yang2022Vtoonify, yang2023styleganex} (see Fig.~\ref{fig:wild}).
Moreover, previous latent optimization methods have limitations in preserving fine-grained details of reference hairstyles (\textit{e.g.}, curl), especially under extreme pose variations.

On the other hand, recent diffusion-based image editing demonstrates superior performance in manipulating human images compared to GANs.
Thanks to the powerful generative prior of text-to-image diffusion models, there has been success in image editing based on human poses~\cite{hu2024animate, kim2024tcan}, depth~\cite{zhang2023adding}, reference images~\cite{yang2023paint}, and clothing~\cite{kim2024stableviton, choi2024improving}.
However, it is still underexplored to leverage diffusion models for hairstyle transfer.

To address the above-mentioned challenges, we propose a novel \textbf{hair}style transfer dif\textbf{fusion} model called \textit{HairFusion}, which generates a high-fidelity image with the reference hairstyle while faithfully preserving the surrounding features such as head shape, clothing, and backgrounds.
We design a one-stage diffusion hairstyle transfer model building upon a pre-trained diffusion model~\cite{rombach2022high} to enhance generalizability on various face images.
Inspired by recent literature on virtual try-on~\cite{choi2021viton, lee2022high, kim2024stableviton}, we posit that hairstyle transfer can be conceptualized as a process of filling in the hair region of a face image, conditioned on the reference hairstyle (\textit{i.e.}, exemplar-based image inpainting). 
Adapting the diffusion-based image inpainting for the hairstyle transfer, we introduce a hair-agnostic representation and a hair align cross-attention (Align-CA).
We first obtain a hair-agnostic representation by thoroughly eliminating the original hair information in the face image.
Next, the Align-CA aligns the reference hairstyle with the hair region of the face image by learning the correspondence between them.
Dense pose representations~\cite{Guler2018DensePose} are provided to the Align-CA to encourage the model to indicate the relative difference in pose and face shape between reference hair and face images.

Nevertheless, when editing the hairstyle via diffusion-based inpainting, challenges remain in effectively preserving essential original features such as identity, clothing, and background of the source face image.
To solve this issue, we introduce adaptive hair blending.
We exploit a cross-attention map to identify hair regions in the output and seamlessly blend them with non-hair areas of the face image during the inference.
Our extensive experiments demonstrate that our method significantly outperforms the existing approaches in terms of synthesized image quality both qualitatively and quantitatively.
Also, our method achieves superior performance than the existing diffusion models for exemplar-based image inpainting, even when applied to diverse, real-world images, as presented in Fig.~\ref{fig:teaser}.
We summarize our contributions as follows:
\begin{itemize}
    \item We present HairFusion, the first one-stage diffusion-based hairstyle transfer framework, capable of generalizing across arbitrary face images in real-world scenarios.
    \item We propose a hair align cross-attention module (Align-CA) that aligns the reference hairstyle with the face image by accounting for differences in head poses. 
     \item Our novel adaptive hair blending effectively preserves the original features in non-hair regions by blending the estimated hair regions of the generated image with the non-hair regions of the face image during inference.
\end{itemize}




\begin{figure*}[t!]
  \centering
  \includegraphics[width=0.9\linewidth]{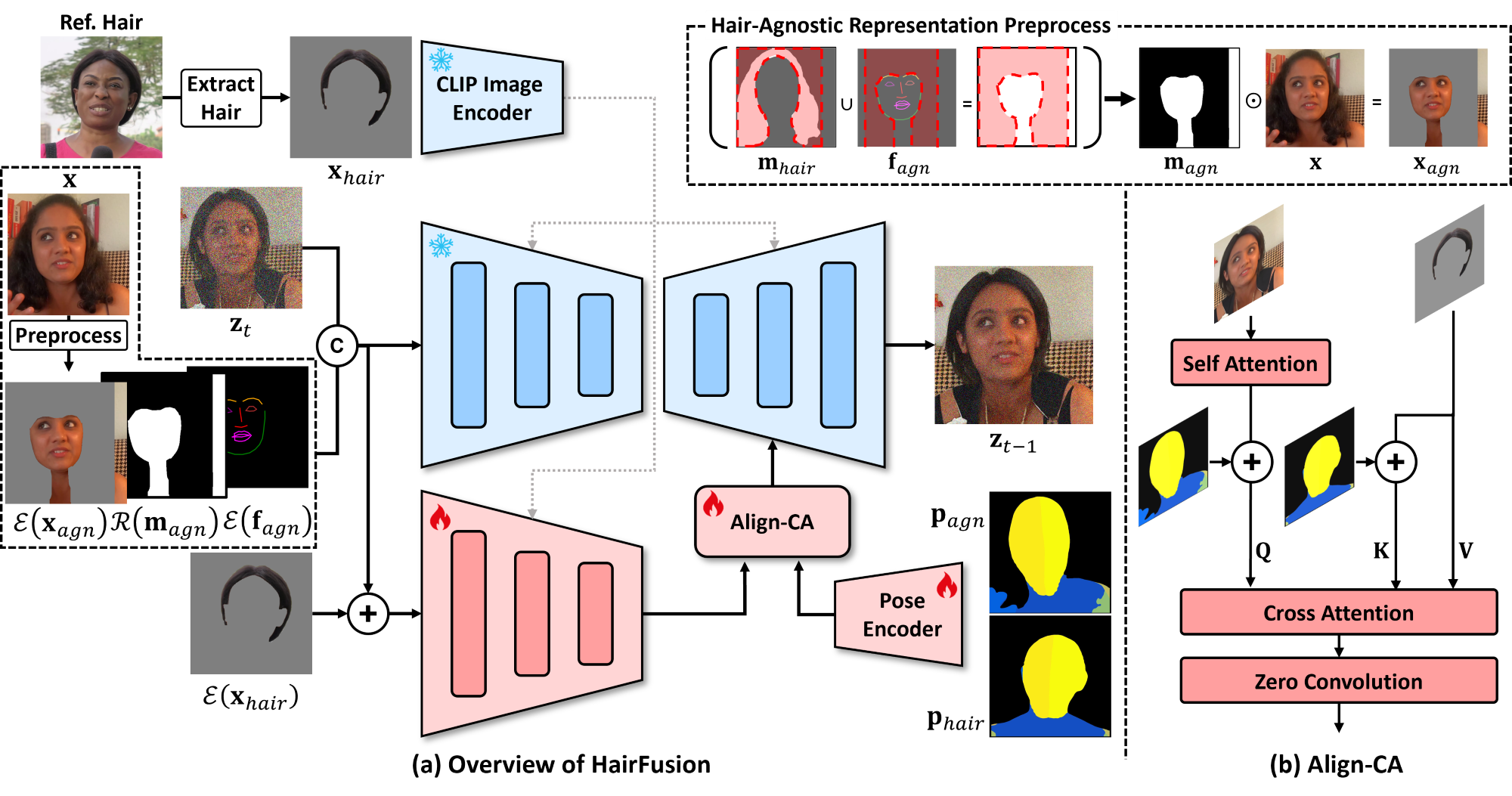}
  \vspace{-0.2cm}
  \caption{Overall pipeline of HairFusion. (a) HairFusion consists of a pre-trained U-Net, a hair encoder, a hair align cross-attention (Align-CA), and a pose encoder. We first preprocess hair-agnostic image $\textbf{x}_{agn}$ using a hair mask $\textbf{m}_{hair}$ and a face outline image $\textbf{f}_{agn}$. Then, we provide $\textbf{x}_{agn}$, a hair-agnostic mask $\textbf{m}_{agn}$, a hair image $\textbf{x}_{hair}$ and dense pose images $\textbf{p}_{agn}$, $\textbf{p}_{hair}$ as inputs to the model. (b) To inject $\textbf{x}_{hair}$ into $\textbf{x}_{agn}$, we leverage the Align-CA, which aligns the hair features with the face features via cross-attention. Here, the pose features are added to the query (Q) and the key (K) as additional guidance.}
  \vspace{-0.4cm}
  \label{fig:overview}
\end{figure*}

\section{Related Work}

\noindent\textbf{Diffusion-based Image Editing.}
Text-to-image diffusion models~\cite{saharia2022photorealistic, rombach2022high} have demonstrated remarkable success in generating high-fidelity images based on textual descriptions.
Several approaches have been proposed for image manipulation under diverse conditions, incorporating conditions such as pose~\cite{hu2024animate, kim2024tcan}, reference images~\cite{yang2023paint, chen2024anydoor}, and multiple conditions~\cite{zhang2023adding, kim2023reference} like depth maps and edge maps.
Furthermore, several studies have extended to image inpainting based on the reference images, including objects~\cite{yang2023paint, chen2024anydoor} or clothing~\cite{kim2024stableviton}.
However, hairstyles, compared to other objects like clothing, involve more diverse inpainting regions due to the high variability in hair width and length across different styles.
This study seeks to broaden the applicability of diffusion models by using them to transfer the hairstyle from the reference image to a source face image.

\noindent\textbf{Hairstyle Transfer.}
The existing hairstyle transfer approaches exploit GANs to apply the reference hairstyle to the face image.
MichiGAN~\cite{tan2020michigan} employed conditional generators incorporating hair attributes such as hair shape and appearance.
LOHO~\cite{saha2021loho} and Barbershop~\cite{zhu2021barbershop} introduced latent optimization techniques based on pre-trained StyleGAN~\cite{karras2020stylegan2} to preserve the overall structure of the face image while reflecting the reference hairstyle.
To broaden the scope of hairstyle transfer applications, HairFIT~\cite{chung2021hairfit}, StyleYourHair~\cite{kim2022style}, StyleGANSalon~\cite{khwanmuang2023stylegan}, and HairFastGAN~\cite{nikolaev2024hairfastgan} have developed the pose-invariant hairstyle transfer, accommodating significant differences in head pose between the face and reference images.
Moreover, HairCLIP~\cite{wei2022hairclip} and HairCLIPv2~\cite{wei2023hairclipv2} interactively edit hairstyles based on multiple conditions, such as text descriptions, reference images, masks, and sketches.
On the other hand, HairNeFR~\cite{chang2023hairnerf} leverages neural
rendering to geometrically align target hair
in the volumetric space.
However, due to the capability of pre-trained StyleGAN, most previous methods still have limitations in generalizing to the face images with various head poses or focal length and preserving fine-grained details of reference hairstyles (\textit{i.e.}, texture and curl). 
In this paper, we introduce a novel diffusion-based hairstyle transfer model performing on multi-view images, enhancing the applicability of hairstyle transfer in real-world scenarios.

\section{Method}

\subsection{Preliminary}
Large-scale diffusion probabilistic models have shown promising performance in image generation.
Diffusion models~\cite{ho2020denoising} learn to generate images from a target data distribution by progressively denoising a normally distributed variable.
Stable Diffusion (SD)~\cite{rombach2022high} is a latent diffusion model (LDM) that performs the denoising process in the latent space of an autoencoder.
To be specific, a pre-trained encoder ($\mathcal{E}$) first transforms an input image $\mathbf{x}$ to latent feature $\mathbf{z}_0 = \mathcal{E}(\mathbf{x})$.
Then, a forward diffusion process is performed with a pre-defined variance schedule $\beta_t$, following denoising diffusion probabilistic models~\cite{ho2020denoising}:
\begin{equation}
    q(\mathbf{z}_t | \mathbf{z}_0) = \mathcal{N}(\mathbf{z}_t ; \sqrt{\bar{\alpha}_t}\mathbf{z}_0, (1-\bar{\alpha}_t)\mathbf{I}),
\end{equation}
where $t \in \{1, ..., T\}$ and $T$ denotes the number of steps in the forward diffusion process.
$\alpha_t$ is defined to be $1-\beta_t$ and $\bar{\alpha}_t$ to be $\Pi_{s=1}^t \alpha_s$. 
SD is trained with the following objective function: 
\begin{equation}
    \mathcal{L}_{LDM} = \mathbb{E}_{\mathcal{E}(\mathbf{x}),\mathbf{y},\epsilon\sim\mathcal{N}(0, 1),t}\left[\lVert\epsilon - \epsilon_{\theta}(\mathbf{z}_t, t, \tau_{\theta}(\mathbf{y}))\rVert_2^2\right],
\end{equation}
where $\epsilon_{\theta}(\cdot)$ is a denoising U-Net~\cite{ronneberger2015u} and $\tau_{\theta}(\cdot)$ is a CLIP~\cite{radford2021learning} text encoder that embeds the text prompt $\mathbf{y}$.


\subsection{Overview}
We address hairstyle transfer from the perspective of exemplar-based image inpainting.
HairFusion aims to newly generate the masked hair region of a source face image $\mathbf{x} \in \mathbb{R}^{3 \times H \times W}$ according to the reference hairstyle $\mathbf{x}_{hair} \in \mathbb{R}^{3 \times H \times W}$.
As presented in Fig.~\ref{fig:overview}(a), we first preprocess a hair-agnostic image $\mathbf{x}_{agn} \in \mathbb{R}^{3 \times H \times W}$ that faithfully removes the original hair and potential hair regions for various reference hairstyles while preserving the regions that need to be reconstructed.
Then, HairFusion takes $\mathbf{x}_{agn}$, $\mathbf{x}_{hair}$, and face outline image $\mathbf{f}_{agn} \in \mathbb{R}^{3 \times H \times W}$ as the input during denoising.
Following Paint-by-Example~\cite{yang2023paint}, we inject the class token of $\mathbf{x}_{hair}$ extracted from a pre-trained CLIP image encoder into the denoising U-Net as an exemplar condition.
Additionally, HairFusion leverages a hair encoder to extract detailed spatial features of $\mathbf{x}_{hair}$.
We align the features of $\mathbf{x}_{hair}$ with $\mathbf{x}_{agn}$ via a hair align cross-attention (Align-CA).
The Align-CA is designed to consider the relative difference of head poses, focal lengths, and face shape between $\mathbf{x}_{agn}$ and $\mathbf{x}_{hair}$ based on their dense pose features.
We obtain dense pose features by injecting $\mathbf{p}_{agn}$ and $\mathbf{p}_{hair}$ into a pose encoder $\mathcal{E}_p$.
During the inference, our newly proposed adaptive hair blending enhances the preservation of the original features in the source face image, which are largely occluded by $\mathbf{m}_{agn}$.

\subsection{Hair-Agnostic Representation}
To apply various reference hairstyles, all the potential hair regions in the source face $\mathbf{x}$ need to be masked as an inpainting region.
Also, the remaining hairstyle information in $\mathbf{x}$ can harm the model’s generalization ability at inference. 
To address this issue, we design a hair-agnostic representation $\mathbf{x}_{agn}$ that eliminates potential hair regions for $\mathbf{x}_{hair}$ and the original hair of $\mathbf{x}$ while preserving the regions that need to be maintained.
As illustrated in the top-right of Fig.~\ref{fig:overview}, we first zero out the hair region of $\mathbf{x}$ according to its hair segmentation mask $\mathbf{m}_{hair}$.
Next, we remove the potential hair region based on the face outline image $\mathbf{f}_{agn}$.
To be specific, we remove the area above the eyebrows and the region extending from the leftmost to the rightmost coordinate along the jawline with additional margins.
We preserve the face region that needs to be accurately reconstructed.
Note that we also preserve the neck and body beneath the chin, where hair typically does not exist.
More details are illustrated in the supplementary material.
Our hair-agnostic representation enables the model to process a wide variety of reference hairstyles while maintaining the core identity features.

\subsection{Hair Align Cross-Attention}
We leverage the cross-attention (CA) to align the reference hairstyle $\mathbf{x}_{hair}$ with $\mathbf{x}_{agn}$.
As shown in Fig.~\ref{fig:overview}(b), our Align-CA exploits reference hairstyle features as the key ($\mathbf{K}$) and value ($\mathbf{V}$), where the output of the followed self-attention layer is given as the query ($\mathbf{Q}$).
We extract intermediate features of the reference hairstyle using a trainable hair encoder whose initial weights are copied from the denoising U-Net.

Since $\mathbf{x}_{hair}$ can vary in face shapes and head poses, it is challenging to accurately determine the hair shape or length based solely on $\mathbf{x}_{hair}$.
For example, even if the hair in $\mathbf{x}_{hair}$ appears to be long, it would be short hair if the reference's face shape is elongated.
Therefore, we add the features of dense pose images, $\mathbf{p}_{agn}$ and $\mathbf{p}_{hair}$, to $\mathbf{Q}$ and $\mathbf{K}$ in Align-CA as follows: 
{\small
\begin{equation}
    \text{Align-CA} = \text{softmax}(\frac{(\mathbf{Q} + \mathcal{E}_p(\mathbf{p}_{agn}))\cdot(\mathbf{K} +\mathcal{E}_p(\mathbf{p}_{hair}))^{T}}{\sqrt{d}})\cdot\mathbf{V},
\end{equation}}
where $\mathcal{E}_p$ is a pose encoder consisting of a stack of convolutional layers and $d$ indicates the dimension of the feature.
In this way, we can assist our model to accurately approximate the hair shape and length in the generated image in relation to the face based on the correspondence between $\mathbf{x}_{hair}$ and $\mathbf{x}_{agn}$.
We replace a few CA layers in the denoising U-Net decoder with our Align-CA.


\subsection{Training Strategy}
For training, we use a multi-view dataset to encourage the model to learn hair alignment across various head poses.
Ideally, training samples consist of pairs with the same identity but different hairstyles and head poses.
Since no existing datasets include such pairs, we utilize a multi-view dataset as an alternative to obtain pairs with the same identity and hairstyle but different head poses.
During the training, one pair is given as $\mathbf{x}$, and the other's hairstyle is given as $\mathbf{x}_{hair}$.

We train the hair encoder, Align-CA, and pose encoder ($\mathcal{E}_p$) of HairFusion with the following objective:
{\small
\begin{equation}
    \mathcal{L}_{LDM} = \mathbb{E}_{\zeta,\epsilon\sim\mathcal{N}(0, 1),t}\left[\lVert\epsilon - \epsilon_{\theta}(\zeta, t, \tau_{\phi}(\mathbf{x}_{hair}), \mathcal{E}(\mathbf{x}_{hair})\rVert_2^2\right],
\end{equation}}
where
{\small
\begin{equation}
    \zeta = (\left[\mathbf{z}_{t};\mathcal{E}(\mathbf{x}_{agn});\mathcal{R}(\mathbf{m}_{agn});\mathcal{E}(\mathbf{f}_{agn})\right], \mathcal{E}_p(\mathbf{p}_{agn}), \mathcal{E}_p(\mathbf{p}_{hair})),
\end{equation}}
$\tau_{\phi}$ is a CLIP image encoder, and $\mathcal{R}$ indicates resize function.

\begin{figure}[t!]
  \centering
  \includegraphics[width=\linewidth]{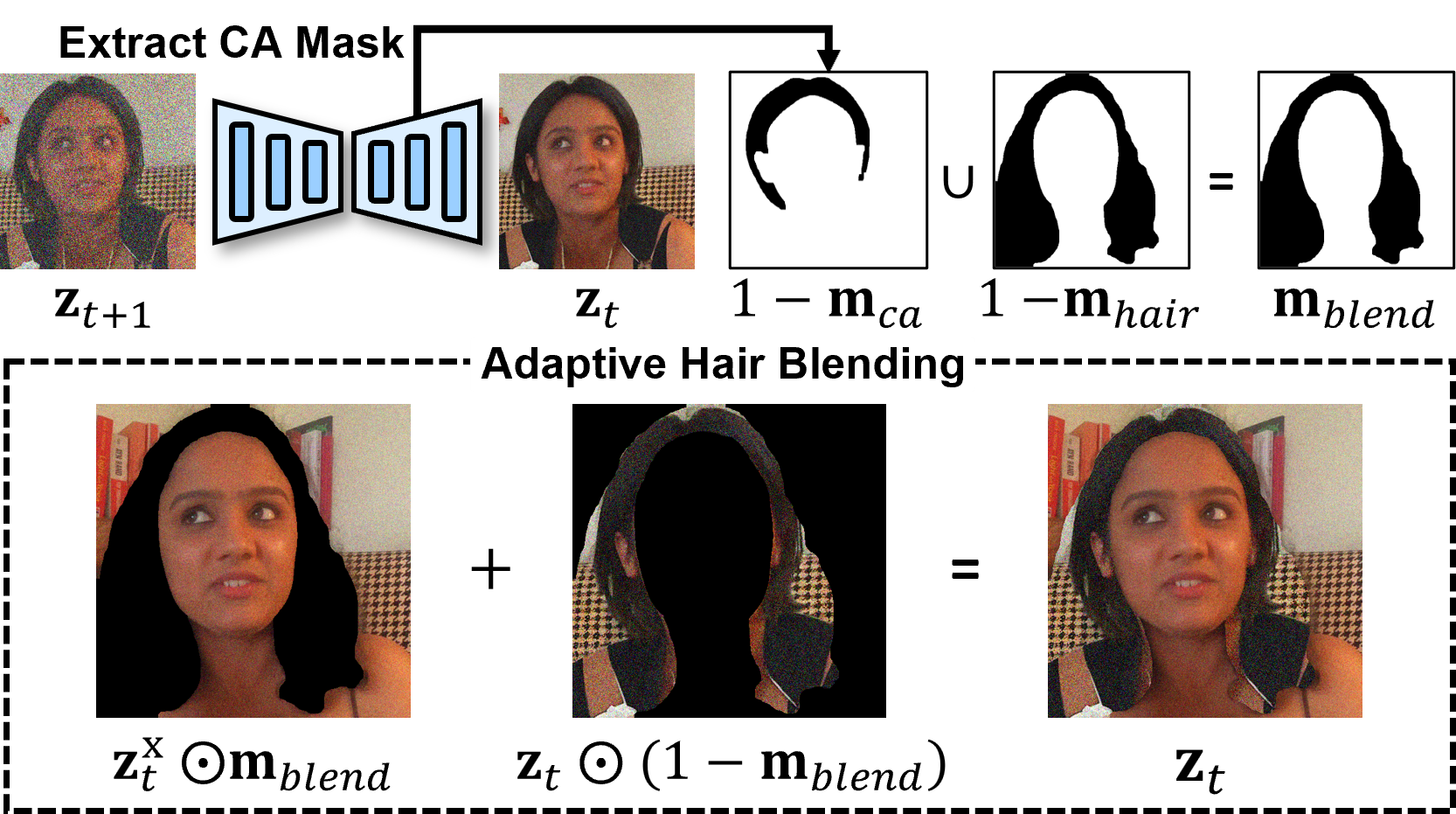}
  \caption{Overview of adaptive hair blending. We obtain $\mathbf{m}_{blend}$ using $\mathbf{m}_{ca}$ extracted from CA maps in Align-CA and the source hair mask $\mathbf{m}_{hair}$. $\mathbf{m}_{blend}$ blends the generated hair features with the other features in the source.}
  \vspace{-0.2cm}
  \label{fig:blending}
\end{figure} 

\begin{figure*}[t!]
  \centering
  \includegraphics[width=\linewidth]{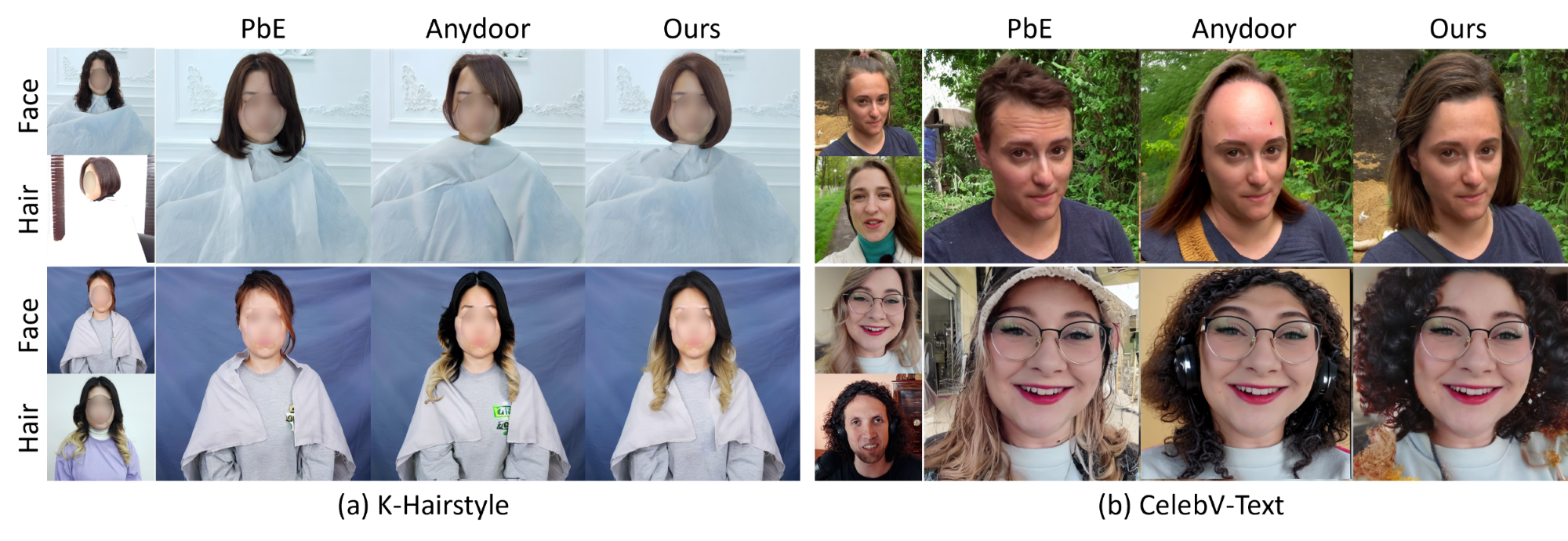}
  \vspace{-0.6cm}
  \caption{Qualitative comparison with the diffusion-based baselines.}
  \vspace{-0.2cm}
  \label{fig:unaligned}
\end{figure*} 

\begin{figure*}[t!]
  \centering
  \includegraphics[width=\linewidth]{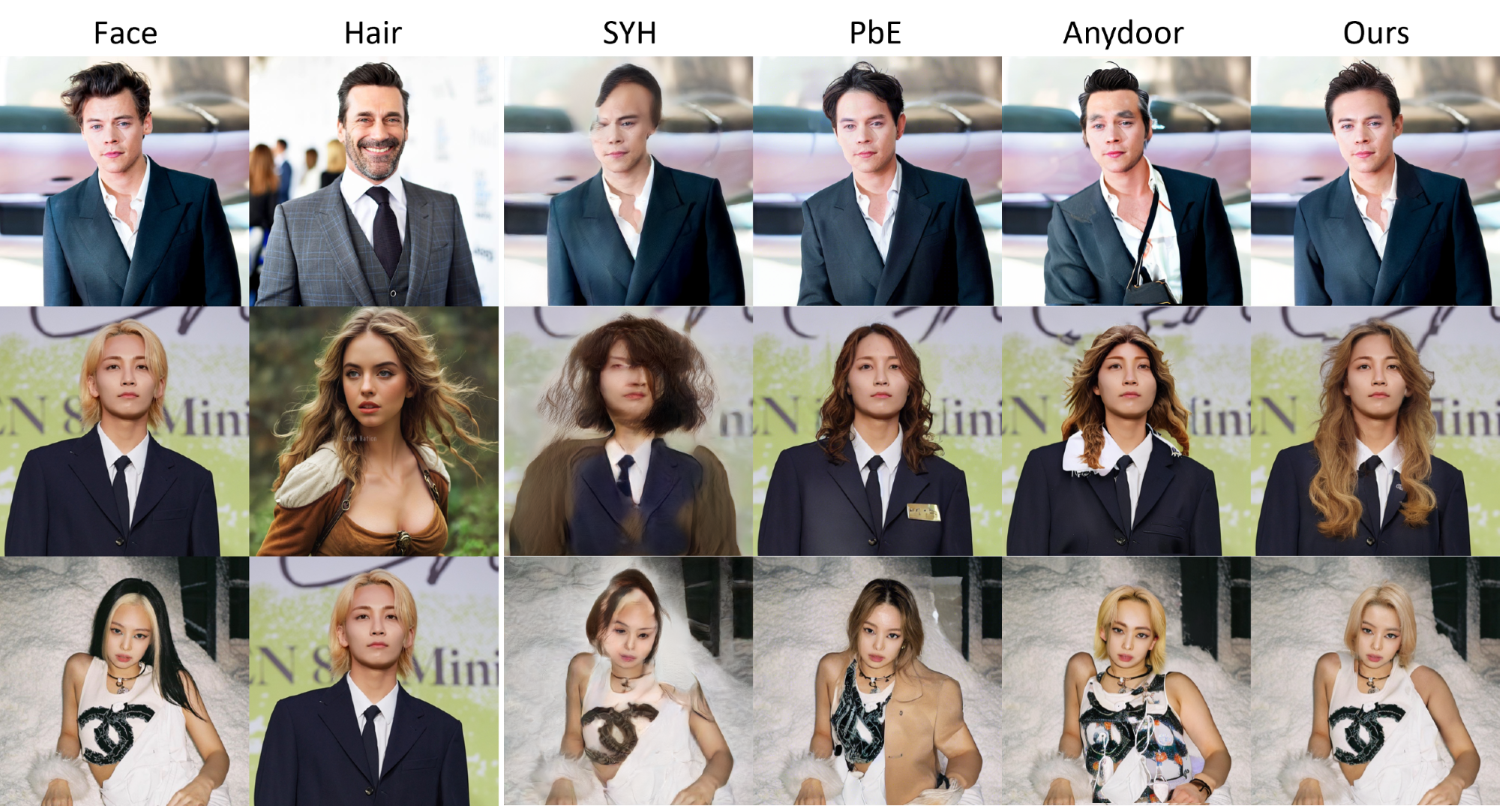}
  \caption{Qualitative comparison with baselines using web-crawled images.}
  \vspace{-0.2cm}
  \label{fig:wild}
\end{figure*} 


\subsection{Adaptive Hair Blending}
\label{sec:adaptive}
One major challenge in hairstyle transfer is to preserve the original features in the source face image except for hair.
However, the occluded region in $\mathbf{x}_{agn}$ inevitably changes due to the autoencoder’s reconstruction error, degrading performance.
As a naive solution, one can apply latent blending as in Blended Diffusion~\cite{avrahami2022blended} using a hair-agnostic mask $\mathbf{m}_{agn}$ as a blending mask.
However, since $\mathbf{m}_{agn}$ is designed to eliminate all the potential hair regions of the generated image, using $\mathbf{m}_{agn}$ as a blending mask unnecessarily loses useful information of $\mathbf{x}$.

To address this issue, we propose an adaptive hair blending that maximally preserves the original features in $\mathbf{x}$.
Since our Align-CA learns to align $\mathbf{x}_{hair}$ with $\mathbf{x}_{agn}$, the aligned hair regions of the generated image are expected to be activated in the CA maps of Align-CA.
With this motivation, we derive an adaptive blending mask ($\mathbf{m}_{blend}$) by leveraging the CA mask ($\mathbf{m}_{ca}$) which indicates the hair region of the output and the source hair mask ($\mathbf{m}_{hair}$) as illustrated in Fig.~\ref{fig:blending}.
Specifically, $\mathbf{m}_{ca}$ is computed by normalizing and thresholding the CA maps extracted from the Align-CA layers.
We obtain $\mathbf{m}_{blend}$ with the union of $1-\mathbf{m}_{ca}$ and $1-\mathbf{m}_{hair}$, indicating the regions to preserve in $\mathbf{x}$ as 1.
Finally, we perform the adaptive hair blending with $\mathbf{m}_{blend}$, and update $\mathbf{z}_{t}$ at timestep $t$ with the blended feature before proceeding to the next timestep ($t-1$) as follows:
\begin{equation}
    \mathbf{z}_{t} \leftarrow \mathbf{z}^{\mathbf{x}}_{t} \odot \mathbf{m}_{blend} + \mathbf{z}_{t} \odot (1-\mathbf{m}_{blend}),
\end{equation}
where $\mathbf{z}^{\mathbf{x}}_{t}$ indicates the noisy latent of $\mathbf{x}$ at timestep $t$, and $\mathbf{m}_{blend}$ is a binary mask, where the regions to be preserved are 1.
We apply adaptive hair blending during the final $N$ timesteps of the denoising process, and $N$ is set as 10 in the entire experiment.
In this way, we can faithfully preserve the features of the source face image $\mathbf{x}$, minimizing the reconstruction error of the autoencoder.

\section{Experiment}
\subsection{Experimental Setup}
\noindent\textbf{Dataset.}
We utilize two multi-view datasets, the K-hairstyle~\cite{kim2021k} and the CelebV-Text~\cite{yu2023celebv} datasets for the experiments.
The K-hairstyle dataset consists of 500,000 high-resolution face images with various focal lengths and head poses with more than 6,400 identities.
Following the Style Your Hair~\cite{kim2022style}, we excluded the images whose hairstyle or face is significantly occluded.
Due to the privacy issue, we blur the face of the images from the K-hairstyle dataset.
Also, the CelebV-Text dataset is a large-scale facial text-video dataset that contains 70,000 in-the-wild face video clips with text pairs.
For the experiments, we randomly sampled 6,000 video clips with different identities and sampled 20 frames from each video.
We remove images whose facial landmarks or hair masks are not detected.
All the images are resized to 512$\times$512 in the experiments.
More details for dataset preprocessing are described in the supplementary material.

\begin{figure*}[t!]
  \centering
  \includegraphics[width=\linewidth]{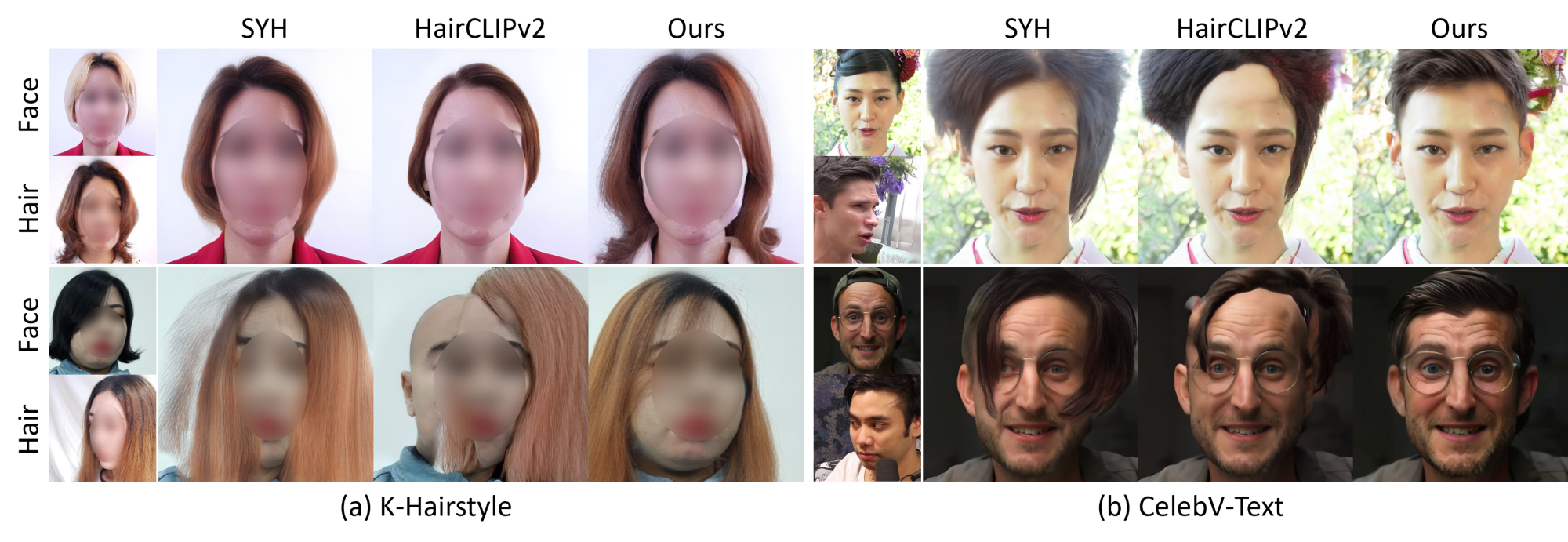}
  \vspace{-0.7cm}
  \caption{Qualitative comparison with StyleGAN-based methods using cropped and aligned images.}
  \vspace{-0.3cm}
  \label{fig:aligned}
\end{figure*} 

\begin{table}[t]
\resizebox{\columnwidth}{!}{%
\begin{tabular}{l|l|c|ccc}
    \hline
    Dataset & Method & FID$\downarrow$ & SSIM$\uparrow$ & PSNR$\uparrow$  & LPIPS$\downarrow$ \\ 
    \hline
    \multirow{4}{*}{K-Hairstyle} 
    & SYH       & 22.84 & 0.62 & 18.41 & 0.32 \\
    & PbE       & 13.26 & 0.59 & 19.26 & 0.29 \\
    & Anydoor   & 19.00 & 0.63 & 18.78 & 0.27 \\
    & Ours      & \textbf{10.82} & \textbf{0.70} & \textbf{21.15} & \textbf{0.18} \\
    \hline
    \multirow{4}{*}{CelebV-Text} 
    & SYH       & 35.77 & 0.62 & 21.14 & 0.30 \\
    & PbE       & 32.72 & 0.41 & 14.55 & 0.42 \\
    & Anydoor   & 35.70 & 0.50 & 15.45 & 0.37 \\
    & Ours      & \textbf{23.50} & \textbf{0.74} & \textbf{22.55} & \textbf{0.18} \\ 
    \hline
\end{tabular}
}
\caption{Quantitative comparison to baselines, including SYH, PbE, and Anydoor.} 
\vspace{-0.3cm}
\label{Tab:unaligned} 
\end{table}

\noindent\textbf{Baselines.}
We first compare our HairFusion with state-of-the-art diffusion models for exemplar-based image inpainting: Paint-by-Example (PbE)~\cite{yang2023paint} and Anydoor~\cite{chen2024anydoor}.
For a fair comparison, we replace their input with our newly designed hair-agnostic representation during the training and inference.
Since Anydoor uses various multi-view datasets to train the model, we fine-tune the Anydoor with our dataset instead of training the model from scratch.
Also, we compare our method with StyleGAN-based hairstyle transfer approaches, Style Your Hair (SYH)~\cite{kim2022style} and HairCLIPv2~\cite{wei2023hairclipv2}.
We apply StyleGAN~\cite{karras2020stylegan2} trained on our dataset to StyleGAN-based approaches.
Since the methods based on StyleGAN pre-trained with the FFHQ dataset~\cite{karras2019stylegan} mainly tackle cropped aligned faces, we also compare our method with StyleGAN-based methods using cropped aligned images.
Specifically, we crop and align the test images in the way the FFHQ dataset is preprocessed for StyleGAN-based approaches. 
Also, we crop and align the results of HairFusion in the same way for comparison.

\noindent\textbf{Evaluation Metric.}
We evaluate our method on two tasks, hairstyle transfer and reconstruction.
Hairstyle transfer modifies hairstyle between two images with different identities and hairstyles.
We measure the fréchet inception distance (FID) score~\cite{heusel2017fid} to evaluate how similar the distributions of the synthesized and the real images are.
In the reconstruction task, we utilize two images with the same identity and hairstyle but different head poses.
One pair is considered the source face (\textit{i.e.}, the ground truth image for the model to reconstruct) while the other serves as the reference hairstyle image.
We measure SSIM, PSNR, and LPIPS for the reconstruction task to evaluate if the model accurately reflects the detailed features (\textit{e.g.}, shape, length, color, etc.) of the reference hair in the result.
For the evaluation, we randomly sample 2,000 pairs from the test sets.

\begin{figure*}[t!]
  \centering
  \includegraphics[width=0.9\linewidth]{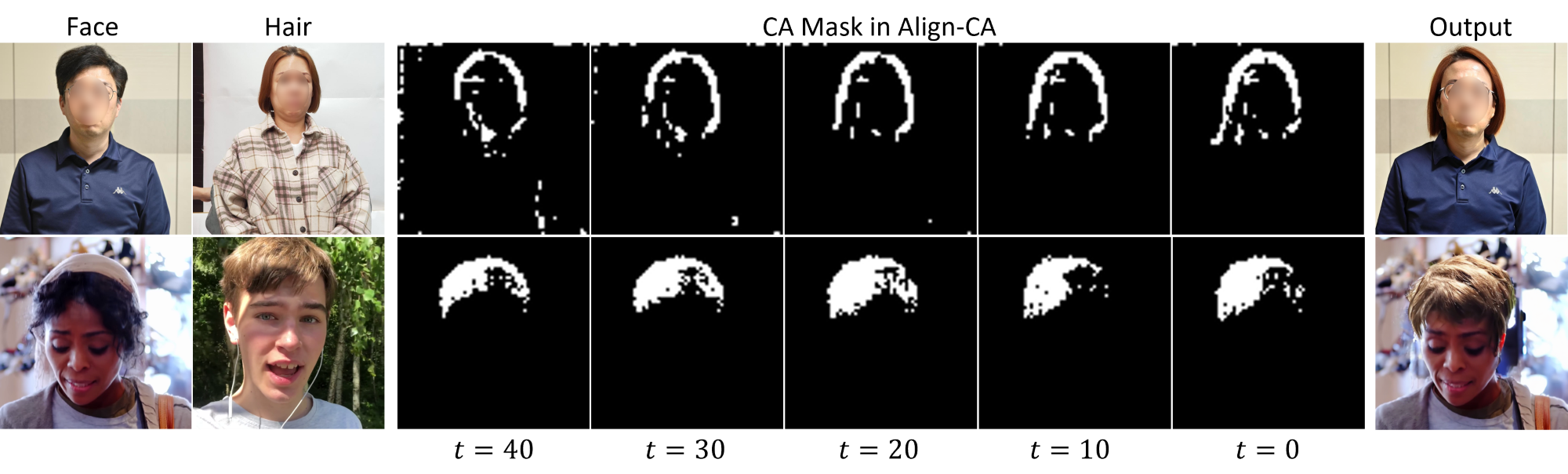}
  \vspace{-0.2cm}
  \caption{Visualization of estimated hair mask in Align-CA. The CA masks are obtained by normalizing and thresholding the CA maps in Align-CA. The CA masks successfully indicate the hair regions of the output. $t$ indicates the timestep of the reverse denoising process, where the total timestep is 50.}
  \vspace{-0.2cm}
  \label{fig:CA}
\end{figure*}

\begin{figure}[t!]
  \centering
  \includegraphics[width=\linewidth]{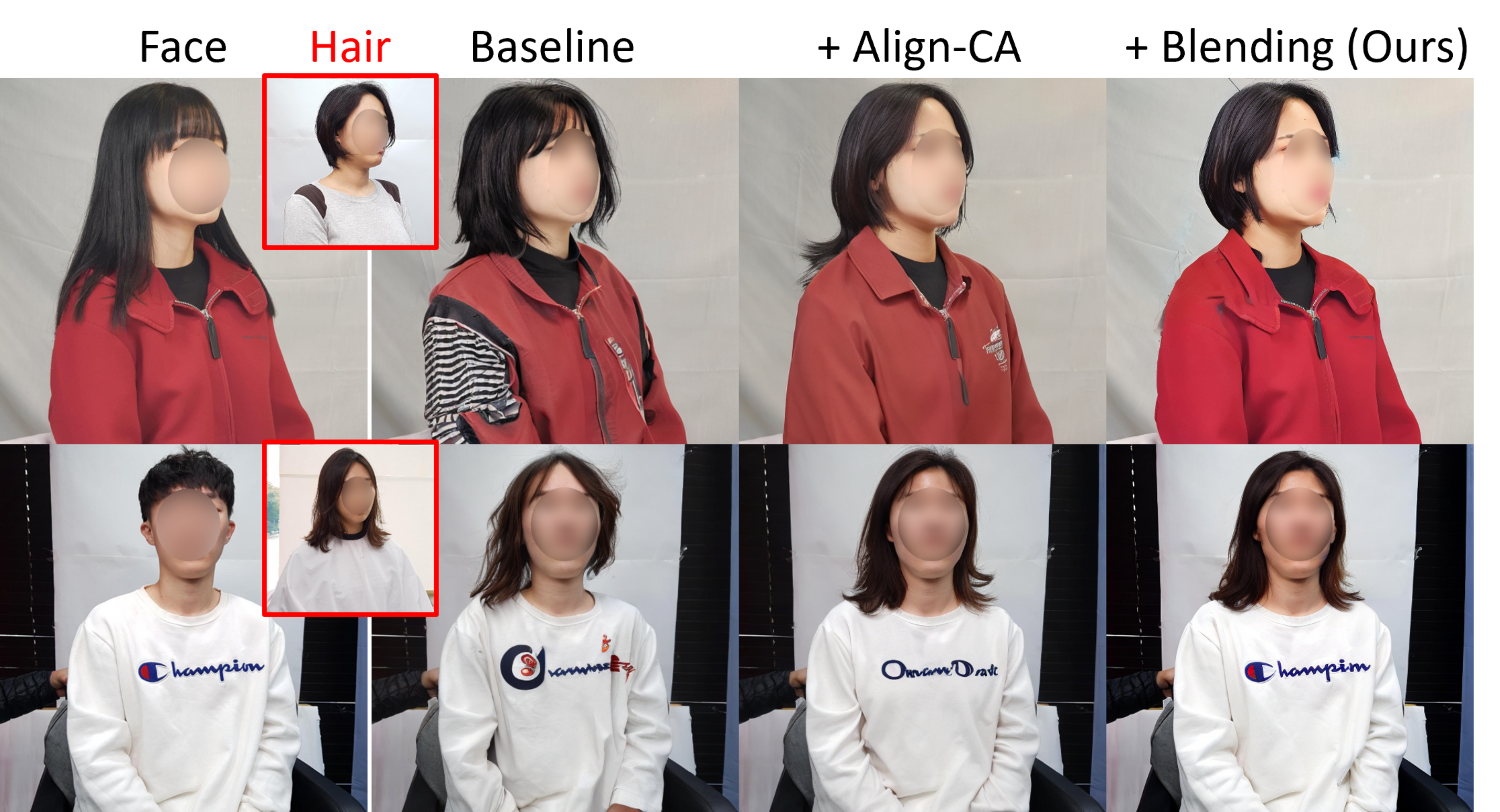}
  \vspace{-0.3cm}
  \caption{Qualitative evaluation for the ablation study using K-Hairstyle dataset. The red box indicates the reference hair.}
  \vspace{-0.3cm}
  \label{fig:abl}
\end{figure}

\subsection{Comparison to Baselines}
\noindent\textbf{Qualitative Comparison.}
Fig.~\ref{fig:unaligned} presents a comparison between our model and diffusion-based image inpainting models, PbE~\cite{yang2023paint} and Anydoor~\cite{chen2024anydoor}, using the K-Hairstyle and CelebV-Text datasets.
PbE struggles to reflect the shape or color of the reference hair in the output.
This is because PbE injects the reference hair features using only the class token extracted by the CLIP image encoder, which results in a significant loss of detailed hair shape and color information.
While Anydoor preserves the features of the reference hair, it fails to maintain the head shape or clothing of the source face image.

Additionally, we conduct a comparison with web-crawled images in Fig.~\ref{fig:wild} using ours and the baselines trained on the K-Hairstyle dataset.
Overall, SYH produces blurry outputs compared to the diffusion-based models.
Although SYH employs StyleGAN trained on multi-view images, it fails to preserve the source face's identity and loses the detail of the reference hair when applied to in-the-wild images containing diverse poses.
Moreover, PbE and Anydoor struggle to maintain both the hairstyle features of the reference hair and the non-hair features in the face image.

Lastly, we compare our method to StyleGAN-based methods using cropped and aligned test sets in Fig.~\ref{fig:aligned}.
The outputs generated by SYH and HairCLIPv2 have blurrier hair compared to ours.
Also, SYH and HairCLIPv2 fail to align the reference hair to a face when they have a large pose difference, even when the inputs are cropped and aligned.

Unlike the existing methods, HairFusion generates realistic outputs not only in both the K-Hairstyle and CelebV-Text datasets but also in the web-crawled in-the-wild images while preserving both the fine details of the reference hair and the surrounding features in the source face image.

\begin{table}[t]
\resizebox{\columnwidth}{!}{%
\begin{tabular}{l|l|c|ccc}
    \hline
    Datset & Method & FID$\downarrow$ & SSIM$\uparrow$ & PSNR$\uparrow$ & LPIPS$\downarrow$ \\ 
    \hline
    \multirow{3}{*}{K-Hairstyle} 
    & SYH        & 25.79 & 0.50 & 15.56 & 0.36 \\
    & HairCLIPv2 & 24.36 & 0.48 & 14.45 & 0.38 \\
    & Ours       & \textbf{15.41} & \textbf{0.55} & \textbf{19.26} & \textbf{0.30} \\ 
    \hline
    \multirow{3}{*}{CelebV-Text} 
    & SYH        & 35.69 & 0.65 & \textbf{22.09} & 0.28 \\
    & HairCLIPv2 & 35.40 & 0.66 & 21.21 & 0.29 \\
    & Ours       & \textbf{22.00} & \textbf{0.70} & 21.77 & \textbf{0.21} \\
    \hline
\end{tabular}
}
\caption{Quantitative comparison to StyleGAN-based baselines using cropped and aligned dataset.} 
\vspace{-0.3cm}
\label{Tab:aligned} 
\end{table}

\noindent\textbf{Quantitative Comparison.}
Table~\ref{Tab:unaligned} shows that our model achieves superior performance compared to the existing methods, including SYH~\cite{kim2022style}, PbE~\cite{yang2023paint}, and Anydoor~\cite{chen2024anydoor}.
For a fair comparison, we apply our newly designed hair-agnostic representation to the diffusion-based baselines.
PbE achieves inferior performance in the reconstruction task (\textit{i.e.}, SSIM, PSNR, and LPIPS), showing that it fails to reflect the detailed features of the reference hair.
Although SYH and Anydoor achieve better reconstruction performance than PbE, they produce unrealistic outputs, achieving higher FID scores.

Also, we compare our method to the existing StyleGAN-based methods in Table~\ref{Tab:aligned} using cropped and aligned images.
Our method achieves superior performance in both hairstyle transfer and the reconstruction task, except for CelebV-Text PSNR.
Since most of the non-hair regions in face images are eliminated in cropped and aligned images, our adaptive hair blending may show only a slight improvement or comparable performance in the reconstruction task.




\subsection{Ablation Study}
We evaluate our method by gradually adding each component of HairFusion to the baseline. 
As in Fig.~\ref{fig:abl} and Table~\ref{Tab:ablation}, we start from the baseline which only contains the denoising U-Net, CLIP image encoder, and the hair encoder.
Then, we gradually add Align-CA and the adaptive hair blending.
The results show that the baseline struggles to estimate the exact hair shape and length of the aligned reference hair, achieving lower reconstruction performance.
Although adding Align-CA largely improves hair alignment performance, it still fails to preserve detailed features in the non-hair region of the face image such as clothing.
Our method with adaptive hair blending achieves the best performance in both hairstyle transfer and reconstruction.
Fig.~\ref{fig:abl} shows that only ours successfully maintains the reference hair features as well as the clothing' details of the face image.


\subsection{Analysis of Adaptive Hair Blending}
Adaptive hair blending leverages the CA mask $\mathbf{m}_{ca}$ in the Align-CA to estimate hair regions of the output.
Fig.~\ref{fig:CA} visualizes $\mathbf{m}_{ca}$ extracted from Align-CA every 10 timesteps as the 50 timesteps of the reverse denoising process proceed.
The first and the second row present a test pair from K-Hairstyle and CelebV-Text, respectively.
We visualize $\mathbf{m}_{ca}$ from the $7$-th CA map in Align-CA while we use the $6$-th and $7$-th CA map in adaptive hair blending.
The figure shows that $\mathbf{m}_{ca}$ successfully indicates the hair regions of the generated image. 
HairFusion effectively reconstructs the original features of $\mathbf{x}$ by blending the generated hair features and the non-hair region features in $\mathbf{x}$.



\begin{table}[t!]
\centering
\resizebox{\columnwidth}{!}{%
\begin{tabular}{l|cccc}
    \hline
    & FID$\downarrow$ & SSIM$\uparrow$ & PSNR$\uparrow$ & LPIPS$\downarrow$ \\ 
    \hline
    Baseline & 15.43 & 0.58 & 18.09 & 0.30 \\
    $+$ Align-CA & 15.59 & 0.62 & 19.33 & 0.27 \\ 
    $+$ Adaptive Hair Blending & \textbf{10.82} & \textbf{0.70} & \textbf{21.15} & \textbf{0.18} \\ 
    \hline
\end{tabular}
}
\caption{Quantitative ablation study using K-Hairstyle.} 
\vspace{-0.2cm}
\label{Tab:ablation} 
\end{table}
\section{Conclusion}
This paper proposes the first one-stage diffusion-based hairstyle transfer model, HairFusion, conceptualizing hairstyle transfer as an exemplar-based image inpainting.
HairFusion introduces Align-CA that aligns the target hairstyle with a face image based on dense pose features, accounting for their pose difference.
Our novel adaptive hair blending technique allows HairFusion to blend the transferred reference hair features with the source face's other appearance and background features based on the hair region estimated by CA maps of Align-CA.
HairFusion achieves state-of-the-art performance compared to the existing approaches, including StyleGAN-based methods and diffusion models for exemplar-based inpainting.
We also demonstrate that HairFusion can generalize to in-the-wild samples with diverse head poses and focal lengths.
\noindent \textbf{\LARGE{Supplementary Material}}
\vspace{0.3cm}
\appendix

\begin{figure*}[t!]
  \centering
  \includegraphics[width=\linewidth]{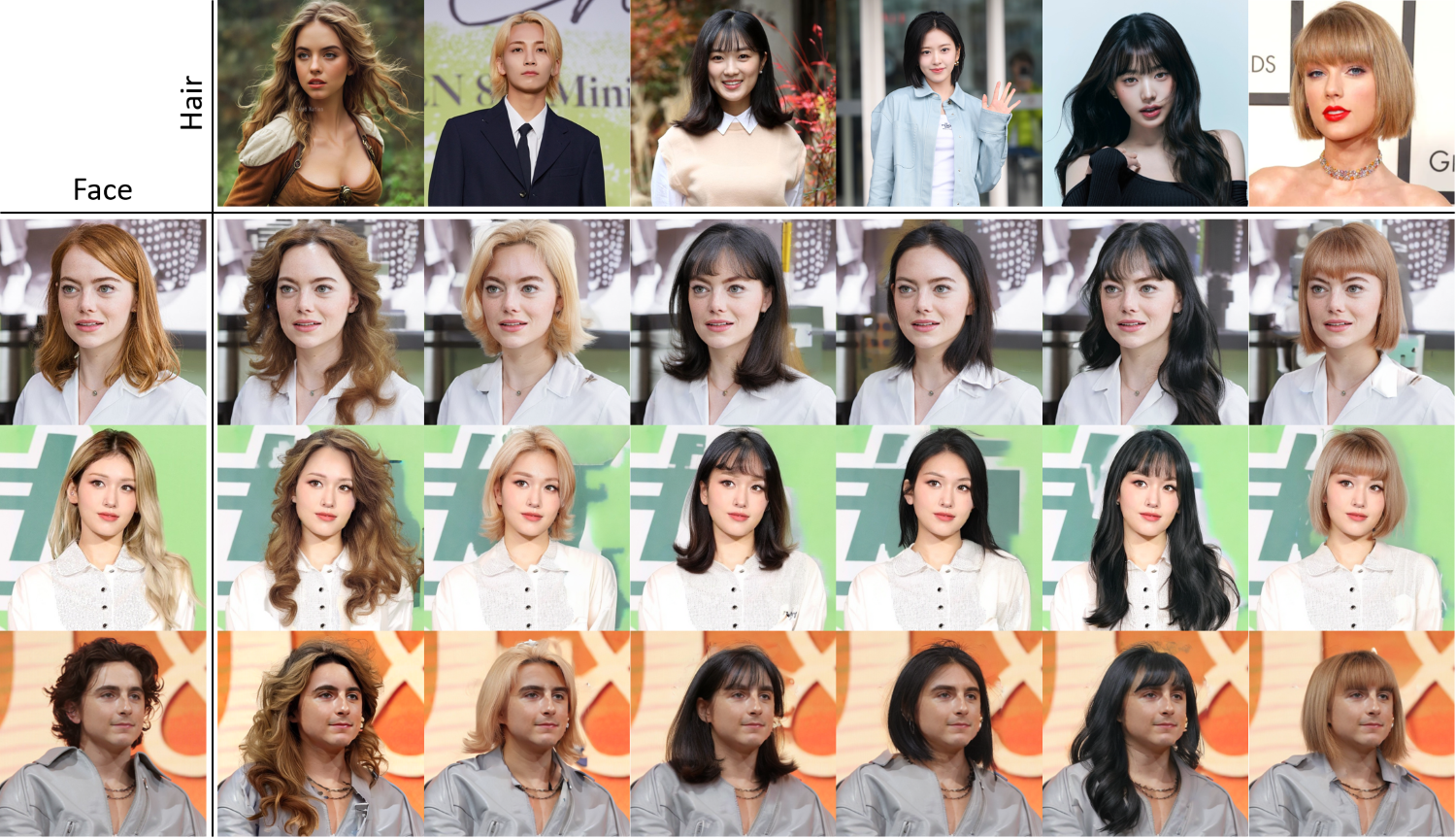}
  \caption{Additional qualitative results of HairFusion using in-the-wild images.}
  \label{fig:grid}
\end{figure*} 

This supplementary material includes a comparison to recent approaches, additional qualitative results using in-the-wild images, details of hair-agnostic representation and dataset, implementation details, limitations, and future work.


\section{Comparison to Recent Approaches}
\begin{table}[h]
\centering
\resizebox{\columnwidth}{!}{%
\begin{tabular}{l|cccc}
    \hline
    & FID$\downarrow$ & SSIM$\uparrow$ & PSNR$\uparrow$ & LPIPS$\downarrow$ \\ 
    \hline
    HairFastGAN & 19.37 & 0.52 & 17.19 & 0.36 \\
    Stable-Hair & 30.02 & 0.50 & 16.91 & 0.32 \\ 
    Ours & \textbf{15.41} & \textbf{0.55} & \textbf{19.26} & \textbf{0.30} \\ 
    \hline		
\end{tabular}
}
\caption{Quantitative comparison to recent approaches using cropped and aligned K-Hairstyle dataset.} 
\label{Tab:recent} 
\end{table}

We conduct a comparison to the most recent StyleGAN-based method, HairFastGAN~\cite{nikolaev2024hairfastgan}, and the concurrent diffusion-based method, Stable-Hair~\cite{zhang2024stablehair}.
We conduct experiments with K-Hairstyle dataset~\cite{kim2021k} using pre-trained models from the official codes and cropped and aligned images as in Table~\ref{Tab:aligned}. 
According to Table~\ref{Tab:recent}, ours achieves superior performance over the others. In our experiments, Stable-Hair may achieve inferior FID due to artifacts generated in bald proxy images. 

\section{Additional Qualitative Results}
Fig.~\ref{fig:grid} shows additional qualitative results of HairFusion using web-crawled in-the-wild images.
The result shows that our method achieves robust performance in real-world scenarios where the images have diverse hair lengths and poses.

\begin{figure*}[]
  \centering
  \includegraphics[width=\textwidth]{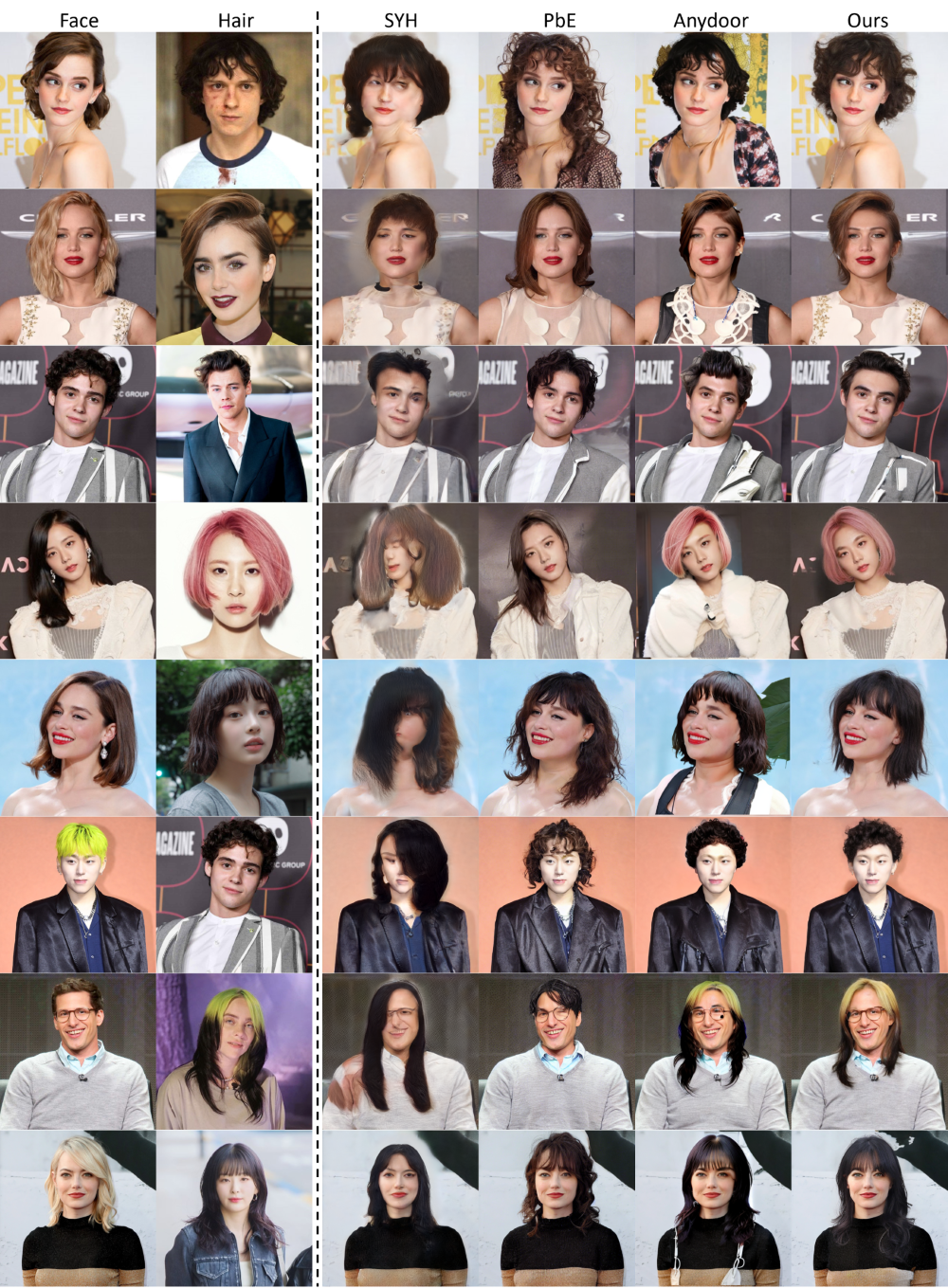}
  \caption{Additional qualitative comparison using in-the-wild images.}
  \label{fig:wild-supp}
\end{figure*}

Moreover, Fig.~\ref{fig:wild-supp} presents additional results of the qualitative comparison to Style Your Hair (SYH)~\cite{kim2022style}, Paint-by-Example (PbE)~\cite{yang2023paint}, and Anydoor~\cite{chen2024anydoor}.
The models are trained with the K-Hairstyle dataset~\cite{kim2021k} that contains multi-view images of various head poses and focal lengths.

According to the figure, SYH generates unrealistic images due to poor hair alignment performance when applied to in-the-wild images of various head poses and focal lengths.
Also, PbE fails to reflect the reference hairstyle including color, shape, and length in the output.
This is because PbE depends solely on the class token of the pre-trained CLIP image encoder to inject hairstyle features, which limits its ability to capture detailed spatial information of the reference hair. 
While Anydoor is better at preserving the reference hairstyle, it generates unrealistic head shapes by simply copying the hairstyle from the reference image (see the fourth and the last rows of Fig.~\ref{fig:wild-supp}).
Moreover, Anydoor struggles to maintain the non-hair region of the source face image such as clothing and background.

In contrast, HairFusion effectively generalizes to web-crawled images by preserving the detailed features of the reference hair through Align-CA, while also maintaining the surrounding features of the source face via adaptive hair blending.

\section{Details of Hair-Agnostic Representation}
Our hair-agnostic representation $\mathbf{x}_{agn}$ is designed to faithfully eliminate potential hair regions for $\mathbf{x}_{hair}$ and the original hair information in $\mathbf{x}$ while preserving the other regions that should be maintained.
We obtain $\mathbf{x}_{agn}$ by masking out $\mathbf{x}$ using the agnostic mask $\mathbf{m}_{agn}$, a binary mask where the regions to be preserved are 1.

To obtain $\mathbf{m}_{agn}$, we first remove the hair region of $\mathbf{x}$ based on its hair mask $\mathbf{m}_{hair}$.
In case $\mathbf{x}$ has no hair, we use DensePose~\cite{Guler2018DensePose} to roughly estimate the head shape and remove the head above the eyebrows as well.
To fully remove the original hair, we eliminate a larger area to account for where hair might potentially be, using the DensePose and facial landmarks.
Specifically, we remove the region from the top of the DensePose to the bottom of the image, and from the leftmost to rightmost jaw points in the facial landmarks with additional margin.
The width of the eliminated region is roughly twice the face width.
We preserve the face region and the neck and body beneath the chin, where hair typically does not exist.

In this way, $\mathbf{x}_{agn}$ minimizes dependency on the original hairstyle during the training, completely obscuring the original hair shape and length.
Furthermore, the removed potential hair region allows various reference hairstyles to be transferred at the inference.

\section{Dataset Details}
We utilize two multi-view datasets, K-Hairstyle~\cite{kim2021k} and CelebV-Text~\cite{yu2023celebv} for the experiments.
In the K-Hairstyle dataset, we use images where the head pose angle (yaw) ranges from -30 to 30 degrees, where 0 degrees indicate a forward-facing position.
We use the ground truth hair mask for the K-Hairstyle and the estimated hair segmentation mask for the CelebV-Text.
For the face parsing masks including the hair mask, we utilize a pre-trained face parsing model~\cite{yu2018bisenet}.
Since the CelebV-Text is a video dataset, we exclude images with extreme motion blur as well.
We generate the face outline image $\mathbf{f}_{agn}$ by drawing lines of different colors based on the facial landmarks extracted by a pre-trained detector~\cite{bulat2017far}.
We obtain dense pose images, $\mathbf{p}_{agn}$ and $\mathbf{p}_{hair}$, using a pre-trained dense human pose estimation model~\cite{Guler2018DensePose}.

\section{Implementation Details}
\subsection{Architecture}
We employ the architecture of the autoencoder and the denoising U-Net of Stable Diffusion v1.4~\cite{rombach2022high}.
We replace the last nine cross-attention layers in the decoder of the denoising U-Net with our Align-CA.
The architecture of the hair encoder follows the encoder of the U-Net.
Also, the pose encoder follows the architecture of the condition embedding network in ControlNet~\cite{zhang2023adding}, which is designed to generate a conditioning vector from a given condition.

\subsection{Training and Inference}
The hair encoder, Align-CA, and pose encoder ($\mathcal{E}_p$) are trainable.
We train our model with the following objective:
{\small
\begin{equation}
    \mathcal{L}_{LDM} = \mathbb{E}_{\zeta,\epsilon\sim\mathcal{N}(0, 1),t}\left[\lVert\epsilon - \epsilon_{\theta}(\zeta, t, \tau_{\phi}(\mathbf{x}_{hair}), \mathcal{E}(\mathbf{x}_{hair})\rVert_2^2\right],
\end{equation}}
{\small
\begin{equation}
    \zeta = (\left[\mathbf{z}_{t};\mathcal{E}(\mathbf{x}_{agn});\mathcal{R}(\mathbf{m}_{agn});\mathcal{E}(\mathbf{f}_{agn})\right], \mathcal{E}_p(\mathbf{p}_{agn}), \mathcal{E}_p(\mathbf{p}_{hair})),
\end{equation}}
where $\mathcal{R}$ indicates resize function.
To further encourage the model to generate a realistic face image, we spatially upweight $\mathcal{L}_{LDM}$ with $(1+\mathbf{m}_{\lambda})$, where $\mathbf{m}_{\lambda} \in \mathbb{R}^{1 \times h \times w}$ is $\mathcal{R}(\lambda_{hair}\mathbf{m}_{hair} + \lambda_{face}\mathbf{m}_{face} + \lambda_{fg}\mathbf{m}_{fg})$. 
Here, $\mathbf{m}_{hair}$, $\mathbf{m}_{face}$, and $\mathbf{m}_{fg}$ denote a hair, face, and foreground mask, respectively.
In the experiment, we set $\lambda_{hair}, \lambda_{face}$, and $\lambda_{fg}$ as 5.
We employ the weights of the autoencoder of Realistic Vision V5.1~\cite{realistic} to enhance the quality of face reconstruction.
The weights of our denoising U-Net are initialized with the pre-trained U-Net of the PbE~\cite{yang2023paint}.
To stabilize the training, we zero-initialize a linear layer after the feed-forward operation in Align-CA and the projection layers followed by the pose encoder.
We train HairFusion using an AdamW optimizer with a fixed learning rate of 1e-4 and a batch size of 48 for 600 and 300 epochs for K-Hairstyle and CelebV-Text, respectively.
We use NVIDIA A100 GPUs for training and NVIDIA RTX A6000 for inference.
For inference, we employ a DDIM sampler~\cite{song2020denoising} with 50 sampling steps.

\section{Limitations and Future Work}
Since HairFusion employs an external face parsing model and facial landmark detector, the output quality highly depends on the performance of the external models.
For instance, in the sixth row of Fig.~\ref{fig:wild-supp}, a single strand of bangs from the reference hair is not reflected in the output because the estimated hair mask failed to capture this detail.
Additionally, HairFusion still struggles to generate or reconstruct detailed features (\textit{e.g.}, text) located near the hair region in the output, as in the third row of Fig.~\ref{fig:wild-supp}.
Lastly, while HairFusion does not currently support additional control over the length or shape of the reference hairstyle, providing users with a preview of the hairstyle in various lengths could significantly improve user satisfaction. Exploring how to control the reference hairstyle based on user input could be a promising direction for future research.

\section*{Acknowledgements}
This work was supported by the Institute of Information \& communications Technology Promotion(IITP) grant funded by the Korea government(MSIT) (No.RS-2019-II190075 Artificial Intelligence Graduate School Program(KAIST) and RS-2021-II212068, Artificial Intelligence Innovation Hub) and the National Research Foundation of Korea(NRF) grant funded by the Korea government(MSIT)(No. 2022R1A5A7083908).

\clearpage
\bibliography{aaai25}

\end{document}